\documentclass[runningheads]{llncs}
\usepackage{graphicx}
\usepackage[T1]{fontenc}
\usepackage{amsmath}

\usepackage{soul}
\usepackage{url}
\usepackage[hidelinks]{hyperref}
\usepackage[utf8]{inputenc}
\usepackage[small]{caption}
\usepackage[usenames,svgnames,hyperref]{xcolor}
\usepackage{booktabs}
\usepackage{tabularx}
\usepackage{algorithm}
\usepackage{algorithmic}
\usepackage{mdframed} 
\usepackage{subcaption}
\usepackage{multirow}
\usepackage{listings}             
\usepackage{multicol} 
\usepackage{float}
\usepackage{xspace}
\usepackage{epstopdf}
\usepackage{tikz}
\usepackage{bm}
\usepackage{forest}
\usepackage[nounderscore]{syntax}

\usepackage{tikz-qtree}

\usepackage[export]{adjustbox}
\usepackage{amssymb}

\graphicspath{{pic/}} 

\usepackage{scalerel}
\usepackage{microtype}
\usepackage{times}

\input pdf-trans
\newbox\qbox
\def\usecolor#1{\csname\string\color@#1\endcsname\space}
\newcommand\outline[1]{\leavevmode%
  \def\maltext{#1}%
  \setbox\qbox=\hbox{\maltext}%
  \boxgs{Q q 2 Tr \thickness\space w 0 0 0 rg 0 G}{}%
  \copy\qbox%
}

\newcommand{\opr}[1]{\ensuremath{\operatorname{\mathit{#1}}}}
\newcommand{\conc}[1]{\ensuremath{\mathsf{#1}}}

\newcommand{\lang}{\mathcal{L}}
\newcommand{\ont}{\ensuremath{\mathcal{O}}\xspace}

\newcommand{\EL}{$\mathcal{EL}^{++}$}
\newcommand{\isa}{\sqsubseteq}

\newcommand{\E}{\exists}

\DeclareMathOperator{\Isa}{\isa}

\DeclareMathAlphabet\mathbfcal{OMS}{cmsy}{b}{n}

\begin{document}

\title{A Survey of Syntactic Modelling Structures\\ in Biomedical Ontologies} 

%
%
\author{Christian Kindermann \and Martin G. Skjæveland}
%
\institute{Department of Informatics, University of Oslo \\
    \email{\{chrikin,martige\}@ifi.uio.no}}
%
%
\maketitle              
\begin{abstract}
    Despite the large-scale uptake of semantic technologies in the biomedical domain,
    little is known about common modelling practices in published ontologies.
    OWL ontologies are often published only in the crude form of sets of axioms leaving the underlying design opaque.
    However, a principled and systematic ontology development life cycle
    is likely to be reflected in regularities of the ontology’s emergent syntactic structure.
    To develop an understanding of this emergent structure,
    we propose to reverse-engineer ontologies taking a syntax-directed approach for
    identifying and analysing regularities for axioms and sets of axioms.
    We survey BioPortal in terms of syntactic modelling trends and common practices for OWL axioms and class frames.  
    Our findings suggest that biomedical ontologies only share simple syntactic structures
    in which OWL constructors are not deeply nested or combined in a complex manner.
    While such simple structures often account for large proportions of axioms in a given ontology,
    many ontologies also contain non-trivial amounts of
    more complex syntactic structures that are not common across ontologies.

\end{abstract}

\setcounter{tocdepth}{2} 
\tableofcontents
\newpage
\section{Introduction}
\label{sec:introduction}

The uptake of OWL in the biomedical domain has lead to
the development of a large number of ontologies as well as
tools providing support for ontology construction and maintenance.  
While some ontologies are documented
to follow pattern-based design principles, e.g., \cite{DBLP:journals/biomedsem/Osumi-Sutherland17,DBLP:journals/biomedsem/SarntivijaiLXMDVSPMPLTSMNBHZSPMCSAH14},
little is known about
what kind of design choices, principles, and patterns are widely-used, 
how they impact ontology engineering in practice.
Comparing ontologies in terms of their design rationales is often challenging
because different ontology are developed and maintained using 
a wide range of methodologies, techniques, and tools. 
Moreover, ontologies are often published as a single file with scarce to no documentation.
Yet, a principled and systematic ontology design is likely to be reflected 
in regularities of the ontology's emergent syntactic structure. 

So, to develop an understanding of common practices in ontology engineering,
we propose to \textit{reverse-engineer} ontologies in terms of syntactic regularities.
Identified regularities may then be analysed and compared to distil
common modelling structures both within and across ontologies.  
In this work, we focus on the syntactic structure of logical expressions in OWL ontologies.
In particular, we analyse the way they are composed and combined.  
The contributions are as follows:
(i) we adapt and simplify the formal framework for identifying syntactic regularities originally proposed in \cite{DBLP:phd/ethos/Kindermann22,kindermann2021syntactic},
(ii) we extend this framework by developing methods for analysing such regularities w.r.t.\ their underlying syntactic structures,
and (iii) we conduct an empirical study to characterise the syntactic structure of axioms and class frames in biomedical ontologies.  

\section{Preliminaries}

We assume the reader to be familiar with Description Logics (DL) \cite{DBLP:conf/dlog/2003handbook}
and the Web Ontology Language (OWL) \cite{DBLP:journals/ws/GrauHMPPS08}.
We use DL notation for the sake of readability but interpret logical constructors as specified by OWL.  
Furthermore, we use both infix and prefix notation for presentational purposes, e.g., $\opr{SubClassOf}(\conc{A},\conc{B})$
may be written as $\conc{A} \isa \conc{B}$ or $\Isa(\conc{A},\conc{B})$.
We disregard OWL annotations, i.e., axioms with and without annotations are indistinguishable.

A directed labelled \textit{graph} $g$ is an ordered pair $(N,E,L)$ where $N$ is a set of nodes,
$L$ is a set of labels,
and $E \subseteq N \times L \times N$ is a set of edges.
A graph $s = (N',E',L')$ is a \textit{subgraph} of $g$, written $s \lesssim g$,
if $N' \subseteq N$ an $E' \subseteq E$.
A \textit{graph isomorphism} between two graphs $g_1 = (N_1,E_1,L_1)$ and $g_2 = (N_2,E_2,L_2)$ is a bijection
$f: N_1 \cup L_1 \rightarrow N_2 \cup L_2$ s.t.\ $(n,l,n') \in E_1$ iff $(f(n),f(l),f(n')) \in E_2$. Two graphs are isomorphic if there exists an isomorphism between them.
A \textit{contraction} of an edge $e = (n_1,l,n_2) \in E$ with $n_1 \neq n_2$ is an operation that first removes $e$ from $E$
and replaces both $n_1$ and $n_2$ with a single node $n'$ and then makes any node (originally) adjacent to either $n_1$ or
$n_2$ adjacent to $n'$. A minor of a graph is a graph obtained by (iteratively) contracting edges, removing edges, or removing nodes
without adjacent nodes.

\section{Framework for Syntax-Directed Analysis of OWL Ontologies}
\label{sec:theory}

\subsection{Syntactic Regularities}
\label{sec:syntacticRegularities}

We analyse structures in OWL ontologies using a syntax-directed approach
based on their abstract representation according to the structural specification for OWL 2 \cite{motik2008}. 
This abstract representation can be captured by abstract syntax trees (AST). 

\begin{definition}[OWL Abstract Syntax Tree]
    \label{def:OWLabstractSyntaxTree}
    Let $\varphi$ be an OWL expression. 
    Then, the \textit{abstract syntax tree} for $T(\varphi)$ is defined as follows:

    \begin{itemize}
        \item if $\varphi$ is atomic, then $T(\varphi)$ is a node labelled with $\varphi$,
        \item if $\varphi = C(\psi_1, \ldots, \psi_n)$, where $C$ is an OWL constructor
            and $\psi_1, \ldots, \psi_n$ are OWL expressions,
            then  $T(\varphi) =$
                    \begin{adjustbox}{valign=t}
                        \begin{forest}
                        for tree={
                          l sep=25pt,
                          parent anchor=south,
                          align=center
                        }
                        [$C$
                            [$T(\psi_1)$,edge label={node[midway,left]{$\ell(\varphi,1)$}} ]
                            [$\ldots$, no edge ]
                            [$T(\psi_n)$,edge label={node[midway,right]{$\ell(\varphi,n)$}} ]
                        ]
                        \end{forest}
                    \end{adjustbox} \\
            where $\ell$ is a labelling function for branches s.t.\ $\ell(\varphi,i)$
            specifies how a subexpression $\psi_i$ at position $i$ is used in relation to $C$.
    \end{itemize} 
\end{definition}
The labelling function $\ell$ is used to treat abstract syntax trees for OWL expressions uniformly as \textit{unordered} trees even in cases where the order of arguments for OWL constructors matters. 
Consider for example the AST of $\opr{SubClassOf}(\conc{A},\conc{B})$.
Here the branches to $\conc{A}$ and $\conc{B}$ would be labelled with "Subclass" and "Superclass" respectively.
In the following, we will not distinguish between OWL axioms and their ASTs, i.e., 
an axiom will be referred to simply as a tree (meaning its AST) and vice versa.
Similarly, an ontology can be understood as a set of trees.

Given the notion of OWL abstract syntax trees,
we can formulate syntax-directed \textit{transformations} for OWL abstract syntax trees
that highlight specific syntactic properties of OWL expressions.
In particular, we can highlight \textit{shared} syntactic properties between OWL axioms
to identify recurring expressions.  
Consider the axioms 
$\alpha_1 = \conc{A_1} \isa \E\,\conc{P}.\conc{A_2}$
and
$\alpha_2 = \conc{B_1} \isa \E\,\conc{Q}.\conc{B_2}$.
While both axioms differ in terms of named classes and properties,
they coincide otherwise.  
This structural similarity can be highlighted via a syntax-directed transformation
that \textit{abstracts} over syntactic properties in which two axioms differ.
For example, with a transformation $G$ that replaces atomic entities with a placeholder symbol, say $*$,
we have $G(\alpha_1) = G(\alpha_2) = * \isa \E *.\,* $~.
Put differently, $\alpha_1$ and $\alpha_2$ exhibit the same syntactic structure that is preserved under the \textit{abstraction} $G$.
An abstraction is intuitively understood as an operation that \textit{hides} some level of detail.
This intuition can be captured for transformations of ASTs by restricting them to the removal of branches and nodes.  
\begin{definition}[Language Abstraction]
    \label{def:languageAbstraction}
    An \emph{abstraction} for a tree language $\mathcal{L}$ 
    into a tree language $\mathcal{L}'$ is defined by a function
    $A \colon \lang \rightarrow \lang'$ such that
  \begin{enumerate}
        \item there exist $t,t' \in \mathcal{L}$ s.t. $t \neq t'$ with $A(t) = A(t')$,
        \item for $t \in \mathcal{L}$ there exists a graph minor $t_m$ that is isomorphic to $A(t)$.  \end{enumerate}
\end{definition}
The second condition formalises the idea of only allowing the removal of a tree's branches and nodes
whereas the first condition requires that an abstraction hides some kind of information so that two syntax trees become \textit{indistinguishable}.  
Coming back to the earlier observation that $G(\alpha_1) = G(\alpha_2)$,
we note that axiom equality under a given abstraction
gives rise to an equivalence relation w.r.t.\ the syntactic structure of axioms in an ontology.
We refer to corresponding equivalence classes as \textit{syntactic regularities}.

\begin{definition}[Syntactic Regularity for Axioms]
    A \emph{syntactic regularity for axioms} in an ontology $\ont$ is an equivalence class $[\alpha]_A =
        \{\alpha_i \in \ont \mid A(\alpha_i) = A(\alpha)\}$, where $A$ is a language abstraction.
\end{definition}

While axioms are the primary building blocks in OWL ontologies,
an entity is often not represented by single axiom but by a \textit{set} of axioms.
So, in addition to regularities for axioms,
we are also interested in regularities for sets of axioms.
We defer the discussion of how to group related axioms
into sets until Section~\ref{ref:modellingStructuresInOWL}.  
Here, we only note that the notion of syntactic regularities for axioms
can be lifted to sets of axioms in a straightforward way.
By abuse of notation, we write $A(S)$ to denote a language abstraction
on \textit{forests} of syntax trees $S$ rather than syntax trees only.

\begin{definition}[Syntactic Regularity for Sets of Axioms]
    Let $\mathcal{S} = \{S_1, \ldots, S_n\}$ be a family of sets of axioms in an ontology $\ont$.
    A \emph{syntactic regularity for sets of axioms} in $\ont$ w.r.t.\ $\mathcal{S}$
    is an equivalence class
    $[S]_A = \{ S_i \in \mathcal{S} \mid A(S) = A(S_i)\}$ where $A$ is a language abstraction.
\end{definition}

\subsection{Modelling Structures}
\label{ref:modellingStructuresInOWL}

A syntactic regularity w.r.t. a language abstraction is uniquely determined
by an abstract syntactic structure, namely the abstract syntax tree or forest
that each of its elements are mapped to under the used language abstraction.  
We will refer to these abstract structures as \textit{modelling structures}. 

\begin{definition}[Modelling Structure] 
    Let $\ont$ be an OWL ontology, $\alpha \in \ont$, and $S \subseteq \ont$, and $A$ a language abstraction.
    Then $A(\alpha)$ and $A(S)$ are \emph{modelling structures} for $\alpha$ and $S$ under $A$ respectively.
\end{definition}  

So, a language abstraction gives rise to syntactic regularities in an ontology
and each syntactic regularity is associated with a modellling structure.
In the following, we provide concrete examples for these notions. 
We already mentioned the language abstraction $G$ 
that highlights structural similarities between axioms by abstracting over atomic entities.
We will refer to this abstraction as the ground generalisation.

\begin{definition}[Ground Generalisation]
    \label{def:abstractionByGroundGeneralisation}
    Let $t$ be an OWL abstract syntax tree.
    The \emph{Ground Generalisation} $G(t)$ of $t$ is a language abstraction defined by a function $G$
    that replaces the label of each leaf node in $t$ with the label $*$~.
  \end{definition}

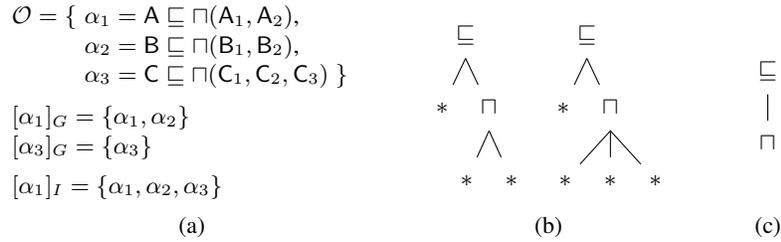
\begin{figure}[t]
  \centering
  \begin{tabular*}{\textwidth}[t]{@{~~~~~~~}c@{~~~~~~}c@{~~~~~~}c@{}}
    \begin{minipage}{0.4\textwidth}
      \begin{tabular}[t]{rl}
        $\ont = \{$ & $\alpha_1 = \conc{A} \isa \sqcap(\conc{A}_1,\conc{A}_2)$, \\
                    & $\alpha_2 = \conc{B} \isa \sqcap(\conc{B}_1,\conc{B}_2)$, \\
                    & $\alpha_3 = \conc{C} \isa \sqcap(\conc{C}_1,\conc{C}_2,\conc{C}_3) \; \}$\\
      \end{tabular}
      
      \vspace{1ex}

      \begin{tabular}[t]{l}
        $[\alpha_1]_G = \{ \alpha_1,  \alpha_2\}$\\
        $[\alpha_3]_G = \{ \alpha_3\}$\\[1ex]
        $[\alpha_1]_I = \{ \alpha_1,  \alpha_2, \alpha_3\}$\\[1ex]
      \end{tabular}
    \end{minipage}
    &
      \begin{minipage}{0.3\textwidth}
        \centering
      \begin{forest}
        for tree={
          l sep=10pt,
          parent anchor=south,
          align=center
        }
        [$\isa$
        [$*$]
        [$\sqcap$
        [$*$ ] 
        [$*$ ]
        ]
        ] 
      \end{forest}
      ~~
      \begin{forest}
        for tree={
          l sep=10pt,
          parent anchor=south,
          align=center
        }
        [$\isa$
        [$*$]
        [$\sqcap$
        [$*$ ] 
        [$*$ ]
        [$*$ ]
        ]
        ] 
      \end{forest}
    \end{minipage}
    &
      \begin{minipage}{0.1\textwidth}
              \centering
        \begin{forest}
          for tree={
            l sep=10pt,
            parent anchor=south,
            align=center
          }
          [$\isa$
          [$\sqcap$] ]
          ] 
        \end{forest}
      \end{minipage}
    \\
    (a) & (b) & (c)\\
  \end{tabular*}
  \caption{Example of the language abstractions $G$ and $I$ applied
    to a sample ontology, and their associated modelling
    structures:
    (a) shows the sample ontology (of three axioms) and its syntactic regularities under
    $G$ and $I$,
    (b) displays the two modelling structures for \ont under $G$,
    while (c) shows the single modelling structure for \ont under
    $I$. Branch labels are not shown.  }
  \label{fig:example}
\end{figure}


The example ontology in Figure~\ref{fig:example}(a)
has two syntactic regularities w.r.t.\ $G$, namely $[\alpha_1]_G = \{\alpha_1, \alpha_2\}$ and $[\alpha_3]_G = \{\alpha_3\}$, which each give rise to a modelling structure under $G$, shown in Figure~\ref{fig:example}(b): 
$G(\alpha_1) = G(\alpha_2) = *\isa\sqcap(*,*)$ and $G(\alpha_3) = *\isa\sqcap(*,*,*)$

Note that all three axioms in the example can be characterised in terms of the nesting of OWL constructors, i.e., all three are subsumption axioms with a conjunction on the right-hand side.  
The nesting of constructors in OWL axioms can be distilled 
with a transformation that removes all leaf nodes (and corresponding branches)
from the axiom's associated abstract syntax tree.
We will refer to the nesting structure of OWL constructors as an axiom's internal tree structure. 

\begin{definition}[Internal Tree Structure]
    \label{def:internalTreeStructure}
    Let $t$ be an OWL abstract syntax tree.
    The \emph{internal tree structure} $I(t)$ of $t$ is a language abstraction defined by
    a function $I$ that removes all leaf nodes and corresponding branches from $t$.  
\end{definition}

The example ontology in Figure~\ref{fig:example}(a)
has only one syntactic regularity w.r.t.\ $I$,
shown in Figure~\ref{fig:example}(c),
since $I(\alpha_1) = I(\alpha_2) = I(\alpha_3)$.
Intuitively, the abstraction $I$ abstracts over more syntactic properties compared to $G$
which leads to fewer but larger syntactic regularities (where the size of a regularity is the number of its elements, i.e., axioms).


As already mentioned in Section~\ref{sec:syntacticRegularities},
conceptual models for domain-specific entities are, more often than not,
represented with a \textit{set of axioms} rather than with a single axiom.
The notion of a \textit{class frame} is widely used for grouping conceptually related axioms in OWL ontologies \cite{DBLP:conf/owled/HorridgeP08,DBLP:journals/dke/NoyMMR04}.

\begin{definition}[Class Frame]
    \label{def:classFrame} 
    A \emph{class frame} $\opr{CF}(\conc{C},\ont)$
    for a class expression $\conc{C}$ in an ontology $\ont$ is defined as the set:
    $\opr{CF}(\conc{C},\ont) = \{ \alpha \in \ont \mid \alpha = \opr{SubClassOf}(\conc{C},\conc{C'}),
    \text{ or } \alpha = \opr{EquivalentClasses}(\conc{C},\conc{C}_1,\ldots,\conc{C}_n)\},
    \text{ or } \alpha = \opr{DisjointClasses}(\conc{C},\conc{C}_1,\ldots,\conc{C}_n)\},$
    $\text{ or }$ $\alpha = \opr{DisjointUnion}(\conc{C},\conc{C}_1,\ldots,\conc{C}_n)\}$.

\end{definition}
The abstractions $I$ and $G$ for abstract syntax trees of axioms
can be lifted to forests of abstract syntax trees in a straightforward manner. 

\begin{definition}[Multiset Lifting of Language Abstractions] 
    Let $F$ be a forest of OWL abstract syntax trees and $A$ a language abstraction for OWL abstract syntax trees.
    Then the image $A(F)$ of $F$ under $A$ is defined as
    the multiset $A(F) = \{A(t) \mid t \in F\}$. 

\end{definition}
We define $A(F)$ as a multiset to account for repetitions of axioms with the same modelling structure.
Consider the set $F = \{\opr{SubClassOf}(\conc{C},\conc{B}), \opr{SubClassOf}(\conc{C},\conc{D})\}$.
Using a set for the lifiting of $G$ would yield 
$\{\opr{SubClassOf}(*,*)\}$ instead of the desired multiset.
We write $\alpha^x$ to denote the $x$-fold repetition of modelling structure $\alpha$. So, $\{\opr{SubClassOf}(*,*)^2\}$ denotes the multiset $\{\opr{SubClassOf}(*,*), \opr{SubClassOf}(*,*)\}$.

%
%


\subsection{Relations between Modelling Structures}
\label{sec:relationsBetweenSyntacticRegularities}

The intention of $G$ with regards to syntactic regularities 
is to group OWL axioms or sets of axioms
based on the way OWL constructors are combined and nested.  
In particular, any difference between axioms
in terms of used OWL constructors will be captured by different syntactic regularities.
Consider the axioms $\alpha_1 = \conc{A} \isa \E\; \conc{R}. \conc{B}$ and 
$\alpha_2 = \conc{A} \isa \E\; \conc{R}.(\E\; \conc{R}. \conc{B})$.
Clearly, $G(\alpha_1) \neq G(\alpha_2)$.
Note, however, that the nesting of OWL constructors in $\alpha_1$, i.e., its internal tree structure $I(\alpha_1)$,
occurs as a \textit{substructure} in $\alpha_2$.
We can formalise this substructure relationship via subgraphs in modelling structures. 

\begin{definition}[Structure Containment] 
    \label{def:internalTreeStructureContainment}
    Let $t$ and $t'$ be two OWL abstract syntax trees.
    Then, $t$ \emph{structurally contains} $t'$, written $t \lesssim_I^G t'$, if
    \begin{enumerate}
        \item $I(t) \lesssim I(t')$ and $I(t) \neq I(t')$, or
        \item $G(t) \lesssim G(t')$ and $I(t) = I(t')$.
    \end{enumerate}
\end{definition}

The two cases in the definition for structure containment are owed to n-ary constructors.
In the case of two OWL expressions $e$ and $e'$ that only involve constructors with a fixed arity
we have that $I(e) = I(e')$ implies $G(e) = G(e)$.
However, this is not the case for expressions involving n-ary constructors.  
Consider for example the axioms
$\alpha_1 = \conc{A} \isa \sqcap(\conc{C}_1, \conc{C}_2)$
and
$\alpha_2 = \conc{A} \isa \sqcap(\conc{C}_1, \conc{C}_2, \conc{C}_3)$.
Here, we have $I(\alpha_1) = I(\alpha_2)$ but $G(\alpha_1) \neq G(\alpha_2)$.
So, defining the substructure containment between OWL abstract syntax trees
only in terms of their internal tree structures would ignore structural information about n-ary constructors.  
The second case in Definition~\ref{def:internalTreeStructureContainment} rectifies this
so that $\alpha_2$ structurally contains $\alpha_1$.
The structure containment relation defines a partial order on OWL abstract syntax trees
and thus induces a partial order on syntactic regularities for axioms.  
 
\begin{lemma}[Partial Order on Ground Generalisations]
    Let $[t_1]_G, \ldots, [t_n]_G$ be syntactic regularities for axioms w.r.t.\ $G$ in an ontology $\ont$.
    Then the relation $\lesssim^G_I$ induces a partial order on $[t_1]_G, \ldots, [t_n]_G$.  
\end{lemma}

Similarly, we can induce a partial order on syntactic regularities for class frames
w.r.t.\ $G$ by defining a containment relation based on a notion of subsets for multisets.
That is, for each number of axioms with the same ground generalisation in one class frame
there needs to exist at least as many axioms with an identical ground generalisation in the other class frame. 

\begin{definition}[Class Frame Containment]
    Let $C$ and $C'$ be class frames in an ontology $\ont$.
    If there exists an injective mapping $m \colon C \rightarrow C'$ 
    s.t.\ $t \in C$ implies that $G(t) = G(m(t))$,
    then $C'$ \emph{contains} $C$, written $C \lesssim_G C'$.
\end{definition}

\begin{lemma}[Partial Order on Class Frames]
    Let $[C_1],\ldots,[C_n]$ be syntactic regularities for class frames in an ontology $\ont$.
    Then the relation $\lesssim_G$ for class frames induces a partial order on
    $[C_1],\ldots,[C_n]$.
\end{lemma}

\section{Methods}

\textbf{Research Questions.} To develop a first understanding of syntactic structures in published ontologies,
we focus on properties related to OWL constructors for class expressions.
In particular, we investigate to what extent such constructors are nested and combined to
give rise to more complex structures.  
Furthermore, we aim to identify and characterise common structures within and across ontologies.
Lastly, we investigate to what extent distinct syntactic structures 
are related by shared substructures.

\label{sec:experimentalDesign}
\textbf{Experimental Design.} 
Since we are interested in the way OWL constructors are used in OWL ontologies,
we will investigate syntactic regularities w.r.t.\ the language abstraction $G$ 
proposed in Section \ref{ref:modellingStructuresInOWL}.
So, we will refer to syntactic regularities based on $G$ (for axioms and class frames)
simply as regularities (for axioms and class frame respectively) unless stated otherwise.
Likewise, we will not explicitly specify that modelling structures for regularities are based on $G$ unless the context is ambiguous.
Our investigation consists of five experiments.
In the following, we give a brief description for each of these experiments
and describe the construction of the experimental corpus of ontologies using BioPortal.  

\textit{1. Number of Syntactic Regularities.}
We determine to what extent ontologies give rise to different regularities, i.e., contain different syntactic structures.

\textit{2. Size of Syntactic Regularities.}
We give an account of the size of syntactic regularities.
Since a regularity is a set, its size is defined by the number of its elements.

\textit{3. Characteristics of Common Modelling Structures.} 
We determine what kind of modelling structures are common within and across ontologies.
For this purpose, we inspect the three largest syntactic regularities in each ontology
and qualify their associated modelling structures in terms of the nesting and combination of OWL constructors.
Furthermore, we compare the modelling structures associated with large regularities across ontologies to identify structures of a general nature. 

\textit{4. Size and Depth of Modelling Structures.} 
We determine to what extent OWL constructors are nested and combined in modelling structures.
For this purpose, we report on the maximal size and depth of modelling structures in ontologies.  
Since a modelling structure for axioms is a tree, its depth is defined as its tree depth,
i.e., the longest path from its root to a child. 
In the case of modelling structures for class frames, their depth is defined as the maximal depth of its axioms.

\textit{5. Interrelations between Syntactic Regularities}.
We determine to what extent syntactic regularities in ontologies are structurally related.
So, we analyse the partially ordered sets of syntactic regularities w.r.t.\ the notions of structural containment (cf.\ Section~\ref{sec:relationsBetweenSyntacticRegularities}). 
In particular, we construct the Hasse diagrams associated with said posets for each ontology
and report on their longest paths, i.e, their depth, as well as their maximal branching factors.

\textbf{Ontology Corpus.} We work with a recent (February 2022) snapshot of BioPortal created in the same way
as described in \cite{matentzoglu_nicolas_2017_439510}. 
The data set of ontologies encompasses a total of 736 ontologies.
We use the OWL API\footnote{\url{http://owlcs.github.io/owlapi/}} (v.5.1.15)
to orchestrate all experiments.
Therefore, we restrict the experimental corpus to ontologies that can be loaded with the OWL API.
We load ontologies without their imports closure to avoid double counting syntactic structures that are imported by different ontologies.  
Furthermore, we exclude ontologies that do not contain class expression axioms
because our experiments are restricted to class expression axioms.
Lastly, we exclude ontologies for which we could not compute all syntactic regularities and their interrelations within one hour.
This procedure results in an experimental corpus of 657 ontologies.

\begin{figure}[t] 
    \include{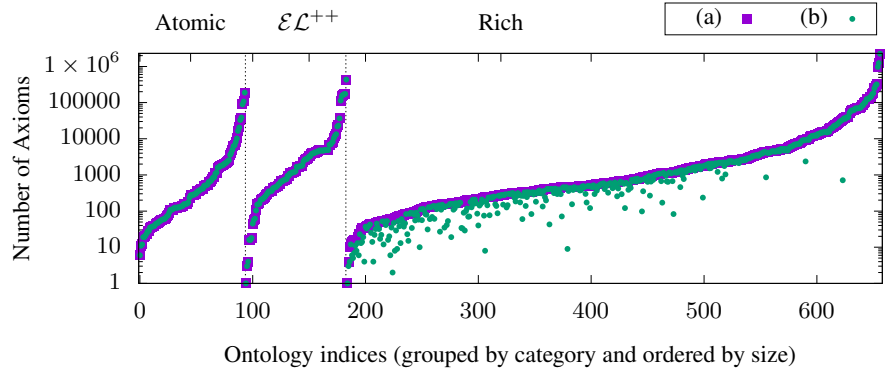}
    \caption{Number of TBox axioms (a) and class expression axioms (b).}
    \label{fig:ontologyTBoxSize}
\end{figure}

In our experiments, we distinguish between three kinds of ontologies.
First, ontologies that consist of atomic axioms only, i.e., \opr{SubClassOf} and 
\opr{EquivalentClasses} axioms that have only named classes as arguments.
Second, ontologies expressible in $\mathcal{EL}^{++}$.
And third, ontologies not expressible in $\mathcal{EL}^{++}$.
We refer to these three kinds of ontologies as atomic, $\mathcal{EL}^{++}$,
and rich ontologies respectively.
Figure \ref{fig:ontologyTBoxSize} shows the size of an ontology’s TBox as well as the size of its subset of class
expression axioms.
We order ontologies within a category by size and assign each ontology an index
in ascending order starting with atomic ontologies as shown in Figure~\ref{fig:ontologyTBoxSize}.
The corpus contains 94 atomic ontologies, 90 $\mathcal{EL}^{++}$ ontologies, and 473 rich ontologies.  

\section{Results}
We present results for the five experiments as specified in Section~\ref{sec:experimentalDesign}
in separate subsections.
We remind the reader that our experimental design distinguishes between three categories of ontologies
(atomic, $\mathcal{EL}^{++}$, and rich) and that we have two experimental conditions
for all three categories, namely, (a) regularities for axioms and (b) regularities for class frames.

\subsection{Experiment 1: Number of Syntactic Regularities}
\label{sec:numberAndSizeOfRegularities}

\begin{figure}[t] 
    \include{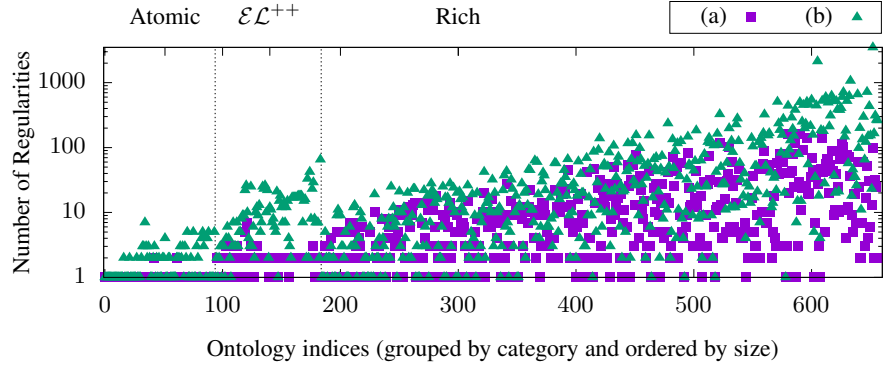}
    \caption{Number of regularities (with respect to $G$) for (a) axioms and (b) class frames in atomic, $\mathcal{EL}^{++}$, and rich ontologies.}
    \label{fig:numberOfRegularities}
\end{figure}

The number of different syntactic regularities for (a) axioms and (b) class frames
are shown in Figure~\ref{fig:numberOfRegularities} for all three categories of ontologies.

The data reveals that atomic and $\mathcal{EL}^{++}$ ontologies give rise to mostly only one or two regularities for axioms whereas rich ontologies give rise to varying numbers of regularities for axioms.
While the largest number of regularities can be found in large rich ontologies, 
it is not the case that all large ontologies give rise to many regularities.

Even though atomic and $\mathcal{EL}^{++}$ ontologies exhibit only a few regularities for axioms
and thus contain mostly axioms of the same syntactic structure,
these axioms are combined in many ontologies to give rise to a comparatively larger number of regularities for class frames.
For example, the $\mathcal{EL}^{++}$ ontology RH-MESH at index 183 has only two regularities for axioms but 65 regularities for class frames.
Similarly, most rich ontologies, especially larger ones beyond index 351 (with about 350 axioms),
often give rise to considerably more regularities for class frames compared to regularities for axioms. For example, the rich ontology FMA at index 652 gives rise to 99 regularities for axioms and 3487 regularities for class frames.

\subsection{Experiment 2: Size of Syntactic Regularities}
The results of Experiment 1 show that many rich ontologies give rise to a fair number of regularities for axioms. In \cite{kindermann2021syntactic}, the same result was found for an older snapshot of BioPortal and it was reported that only a few of these regularities for axioms are large. 
In particular, in the case of regularities for axioms,  it was determined that 90\% of axioms in many ontologies can be covered by one to three regularities in all three ontology categories.
However, the same could not be reported for regularities of class frames; especially for larger rich ontologies.
In the case of class frames, it was reported that often more than ten regularities are required to account for 90\% of axioms in a given ontology.

\begin{table}[t]
    \centering
    \caption{Number of ontologies giving rise to a minimal number of regularities (both for axioms and class frames) with a minimal size of 10, 100, and 1000.}
\label{table:regularitySize}

\begin{tabular}{| c | r | r | r | r || r | r | r | r | }
\hline
\multirow{3}{*}{\textbf{Min. Regularities}} & \multirow{3}{*}{\textbf{Min. Size}} & \multicolumn{6}{c|}{\textbf{Number of Ontologies}}\\
\cline{3-8}
 & & \multicolumn{3}{c||}{Regularities for Axioms} & \multicolumn{3}{c|}{Regularities for Class Frames}\\
\cline{3-8}
 &
 & \multicolumn{1}{c|}{Atomic}
 & \multicolumn{1}{c|}{~~~$\mathcal{EL}^{++}$~~~}
 & \multicolumn{1}{c||}{~~~~Rich~~~~}
 & \multicolumn{1}{c|}{Atomic}
 & \multicolumn{1}{c|}{~~~$\mathcal{EL}^{++}$~~~}
 & \multicolumn{1}{c|}{Rich}
 \\
 \hline
 \multirow{3}{*}{5}   & 10   & - & - & 127 & 1 & 37 & 189\\
                      & 100  & - & - & 35  & 1 & 6 & 64\\
                      & 1000 & - & - & 8   & - & 1 & 21\\
 \hline
 \multirow{3}{*}{10}  & 10   & - & - & 45 & - & 9 & 108\\
                      & 100  & - & - & 7  & - & 1 & 34\\
                      & 1000 & - & - & -  & - & - & 5\\
 \hline
\end{tabular} 
\end{table}

While this finding gives some indication for the size of the three largest regularities in ontologies,
it is important to keep in mind that many ontologies in our experimental corpus contain several thousands of axioms
and that small relative proportions of an ontology can still correspond to many axioms. 
So, to give an account of the size of regularities in terms of absolute numbers,
we report on the number of ontologies that contain at least five or ten regularities with a minimal size of (i) ten, (ii) a hundred or (iii) a thousand elements in Table~\ref{table:regularitySize}. 

It transpires that mostly rich ontologies give rise to multiple regularities of non-trivial sizes within a given ontology.
In the case of regularities for axioms, for example, there are 35 rich ontologies with at least 5 regularities that have at least 100 elements.  
In the case of regularities for class frames, there are even 34 rich ontologies with at least 10 regularities that have at least 100 elements.  
This confirms to some extent the hypothesis that there exist ontologies with more than three regularities of non-trivial size.
However, increasing either the number of minimal regularities, e.g., to ten, or the number of minimal elements, e.g., to 1000,
reveals that there are only a few ontologies with many regularities of considerable size.

Lastly, we note that many rich ontologies \textit{do not} give rise to at least 5 regularities with a minimal size of ten.
This is interesting in the context of the total number of ontologies (cf.\ Section~\ref{sec:numberAndSizeOfRegularities})
that give rise to 5 or more regularities.
In the case of regularities for axioms, there are 285 such rich ontologies
which means that $285 - 127 = 158$ ontologies contain only a few large regularities despite giving rise to 5 or more.
Similarly, in the case of regularities for class frames, there are $364 - 189 = 175$ such ontologies.

\subsection{Experiment 3: Characteristics of Common Modelling Structures}
\label{sec:experiment3}
We remind the reader that each syntactic regularity is associated with a unique modelling structure.
So, we can identify common syntactic structures \textit{within} an ontology by inspecting the modelling structures 
of the ontology's largest regularities.
Furthermore, we can identify common syntactic structures \textit{across} ontologies by comparing modelling structures
associated with the largest regularities within ontologies.

The three largest regularities for axioms across atomic, $\mathcal{EL}^{++}$, and rich ontologies 
give rise to 2, 11, and 103 distinct modelling structures respectively.
Table~\ref{table:commonStructuresAxioms} lists those modelling structures\footnote{The prefix ``Object'' in some OWL expressions is abbreviated with the capital letter ``O'' for presentational purposes.}
that occur across at least 20 different ontologies. 
The values in the last three columns of Table~\ref{table:commonStructuresAxioms}
reveal the actual number of ontologies 
in which a given modelling structure is associated with one of the three largest regularities,
e.g., the modelling structure $\opr{EquivalentClasses}(*,*)$ is associated with one of the three largest regularities in two atomic ontologies, two $\mathcal{EL}^{++}$ ontologies, and 24 rich ontologies. 

Overall, it transpires that only a few modelling structures for axioms are common both within and across ontologies.
Furthermore, these modelling structures are fairly simple in regards to the way OWL constructors are nested and combined. 
Nevertheless, it is important to keep in mind that rich ontologies
exhibit a large variety of modelling structures that are associated with their respective largest regularities. It is also important to mention that many such structures are more complex compared to the ones shown in Table~\ref{table:commonStructuresAxioms}.
For example, the second largest regularity in the ontology HOOM with 78738 elements is associated with the modelling structure
\begin{center}
\opr{EquivalentClasses}(*, \opr{ObjectIntersectionOf}(\opr{ObjectSomeValuesFrom}(*,*), \opr{ObjectSomeValuesFrom}(*,*), \opr{ObjectSomeValuesFrom}(*,*), \opr{ObjectSomeValuesFrom}(*,*), \opr{DataHasValue}(*,*))).
\end{center}
So, while common modelling structures for axioms \textit{across} ontologies are mostly simple,
common modelling structures \textit{within} ontologies can also be rather complex.

\begin{table}[t]
    \centering
    \caption{Common modelling structures across ontologies. A modelling structure is considered common in a given ontology if it associated with one of its three largest regularities. Ordered by total number of ontologies.}
\label{table:commonStructuresAxioms}
\resizebox{\textwidth}{!}{%
\begin{tabular}{| c | l | r | r | r |}
\hline
Row & \multicolumn{1}{c|}{Modelling Structure} & Atomic & $\mathcal{EL}^{++}$ & Rich \\
\hline
1 & $\opr{SubClassOf}(*,*)$ & 94 & 88 & 466\\
2 & $\opr{SubClassOf}(*,\opr{OSomeValuesFrom}(*,*))$ & - & 68 & 270\\
3 & $\opr{DisjointClasses}(*,*)$ & - & - & 103\\
4 & $\opr{EquivalentClasses}(*, \opr{OIntersectionOf}(*, \opr{OSomeValuesFrom}(*,*)))$ & - & 1 & 70\\
5 & $\opr{SubClassOf}(*, \opr{OAllValuesFrom}(*,*))$ & - & - & 44\\
6 & $\opr{EquivalentClasses}(*,*)$ & 2 & 2 & 24\\
7 & $\opr{SubClassOf}(*,\opr{OExactCardinality}(*,*,*))$ & - & - & 20\\
 \hline
\end{tabular} }
\end{table}

The three largest regularities for class frames across atomic, $\mathcal{EL}^{++}$, and rich ontologies give rise to 6, 28, and 209 distinct modelling structures respectively.
Table~\ref{table:commonStructuresClassFrames} lists those modelling structures for class frames
that occur across at least 20 different ontologies in the same manner as Table~\ref{table:commonStructuresAxioms} lists modelling structures for axioms.
The results are similar to the case for regularities for axioms
in the sense that common modelling structures for class frames across ontologies are mostly simple, i.e.,
the class frames consist of only a few axioms and the axioms are not deeply nested. 
Likewise, there are also many ontologies in which the largest three regularities for class frames  
are associated with more complex modelling structures involving more axioms or more deeply nested OWL constructors (see regularities in CLO for example). 
However, such more complex modelling structures are only common within ontologies and not across.

\begin{table}[t]
    \centering
    \caption{Number of ontologies in which its the three largest regularities for class frames is associated with a given modelling structure. Ordered by total number of ontologies.}
\label{table:commonStructuresClassFrames}

\resizebox{\textwidth}{!}{%
\begin{tabular}{| c | l | r | r | r | }
\hline
Row & \multicolumn{1}{c|}{Modelling Structure} & Atomic & $\mathcal{EL}^{++}$ & Rich \\
\hline
1 & $\{\opr{SubClassOf}(*,*)^1\}$ & 94 & 67 & 431\\
\hline
2 & $\{\opr{SubClassOf}(*,*)^2\}$& 37 & 32 & 106\\
\hline
3 & $\{\opr{SubClassOf}(*,*)^3 \}$& 16 & 15 & 20\\
\hline
4 & $\{\opr{SubClassOf}(*,\opr{OSomeValuesFrom}(*,*))^1 \}$& - & 22 & 12\\
\hline 
5 & $\{\opr{EquivalentClasses}(*, OIntersectionOf(*, \opr{OSomeValuesFrom}(*,*)))^1 \}$& - & 1 & 62\\
\hline
6 & $\{\opr{DisjointClasses}(*,*)^1 \}$& - & - & 22\\
\hline
7 & $\{\opr{SubClassOf}(*,*)^1, \opr{SubClassOf}(*,\opr{OSomeValuesFrom}(*,*))^1 \}$& - & 35 & 157\\
\hline
8 & $\{\opr{SubClassOf}(*,*)^1, \opr{SubClassOf}(*,\opr{OSomeValuesFrom}(*,*))^2 \}$& - & 9 & 50\\
\hline
9 & $\{\opr{SubClassOf}(*,*)^1, \opr{SubClassOf}(*,\opr{OSomeValuesFrom}(*,*))^3 \}$& - & 15 & 8\\
\hline
10 & $\{\opr{SubClassOf}(*,*)^1, \opr{DisjointClasses}(*,*)^1 \}$& - & - & 58\\
\hline
11 & $\{\opr{SubClassOf}(*,*)^1,\opr{EquivalentClasses}(*,*)^1 \}$& 2 & - & 19\\
\hline
\end{tabular} }
\end{table}

\subsection{Experiment 4: Size and Depth of Modelling Structures}
\label{sec:experiment4}
\begin{figure}[t] 
    \include{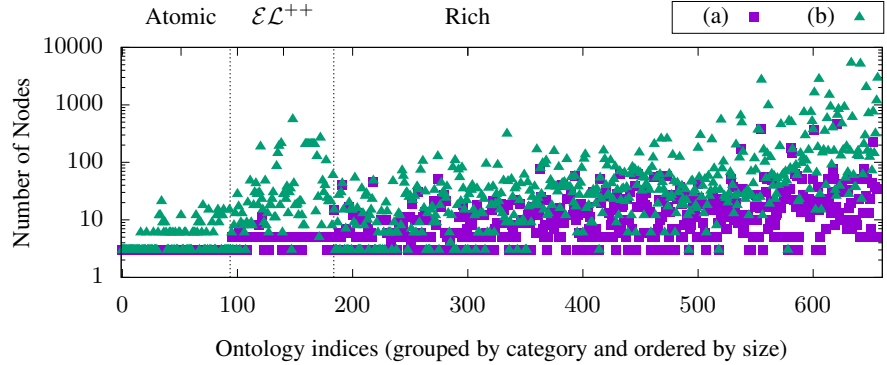}
    \caption{Number of nodes in the largest modelling structures associated with regularities for (a) axioms and (b) class frames.}
    \label{fig:structureSize}
\end{figure}

In this section, we shed some light on the most complex modelling structures in ontologies.
We start with the size of modelling structures, i.e., their number of nodes. 
Figure~\ref{fig:structureSize} shows the size of the largest modelling structures in ontologies
for both (a) axioms and (b) class frames.
We will first highlight some details about the size of modelling structures for axioms
before we compare them to modelling structures for class frames.  

The maximal size of modelling structures for axioms in atomic ontologies is three because they 
only contain the modelling structures $* \isa *$ and $* \equiv *$~.
Similarly, the size of modelling structures in most \EL ontologies is three or five because
they only contain the modelling structures $* \isa *$  and $* \isa \E *.\,*$~.
There are only four ontologies containing modelling structures with a size larger than five.
The largest one is found in the ontology CHIRO with size 11 and has the form
$* \equiv * \sqcap (\E *.(* \sqcap (\E *.*)))$.
However, about half of rich ontologies (211 out of 473) contain modelling structures for axioms
with a size larger than ten.
Interestingly, the maximal size of modelling structures in ontologies appears be independent of the ontologies' overall size, i.e., modelling structures of different sizes occur in ontologies of different sizes.

The maximal size of modelling structures for class frames is often considerably larger
compared to the maximal size of modelling structures for axioms, especially for \EL and rich ontologies that have more than about 350 axioms. This is to be expected if class frames consist of combinations of many axioms. 
In this regard, it transpires that class frames in many atomic ontologies and many rich ontologies of smaller size  
consist of only single axioms.
On the right-hand side of Table~\ref{table:structureDepth},
we summarise how many ontologies contain class frames up to a maximal number axioms.
It appears that \EL and rich ontologies contain class frames with more than three axioms
whereas many atomic ontologies only contain class frames with one or two axioms.  

In addition to the size of modelling structures, we also investigate their depth.  
Note that the depth of a class frame is defined in terms of the maximal depth of its axioms.
So, the maximal depth of modelling structures for both axioms and class frames is the same and we will not distinguish between the two in the following. 
On the left-hand side of Table~\ref{table:structureDepth},
we summarise how many ontologies contain modelling structures up to a maximal depth.
There are 167 rich ontologies that contain modelling structures with a depth of at least four.
This shows that many rich ontologies not only contain fairly large modelling structures but that
 modelling structures also involve non-trivial nestings of OWL constructors.

\begin{table}[b]
    \centering
    \caption{Maximal nesting depth of modelling structures (left-hand side) and maximal number of axioms in class frames (right-hand side).}
\label{table:structureDepth}
\begin{tabular}{| r | r | r | r | r | }
\hline
\textbf{Max Depth} & Atomic & $\mathcal{EL}^{++}$ & Rich \\
 \hline
 1      & 94 & 16 & 107\\
 2      & -  & 71 & 116\\
 3      & -  & 1  & 83\\
 4      & -  & 1  & 36\\
 5--9   & -  & 1  & 118\\
 $\ge$ 10 & -  & -  & 13\\
 \hline
\end{tabular} 
    \hspace*{1cm}
\begin{tabular}{| r | r | r | r | r | }
\hline
\textbf{Max CF Axioms} & Atomic & $\mathcal{EL}^{++}$ & Rich \\
 \hline
 1 & 54 & 7 & 43\\
 2 & 21 & 14 & 46\\
 3 & 6& 11 & 49\\
 4--9 & 13 & 35 & 174\\
 10--19 & - & 5 & 73\\
 $\ge$ 20 & - & 18 & 88\\
 \hline
\end{tabular} 
\end{table}

\subsection{Experiment 5: Interrelations between Syntactic Regularities}

Table~\ref{table:hierarchyDepthBranching} shows the depth and maximal branching factor
of Hasse diagrams corresponding to partially ordered sets for syntactic regularities for axioms and class frames w.r.t.\ $\lesssim_I^G$ and $\lesssim_G$ respectively.  
It transpires that more than half of the ontologies in our experimental corpus (365 out of 657) give rise to Hasse diagrams with a depth of at least 4. Moreover, 110 ontologies even bring about Hasse diagrams with a depth of 10 or more. The numbers for the maximal branching factor are comparable.

A long path in a Hasse diagram for regularities of class frames means that corresponding modelling structures for class frames are based on the same constituent components since $\lesssim_G$ is defined in terms of a subset relation for multisets.
A large branching factor, on the one hand, means that
many class frames share a common substructure, namely the modelling structure of their parent.
But, on the other hand, it also means that siblings of that parent vary in terms of the modelling structures.

Similarly, a long path in a Hasse diagram for regularities for axioms (as in the case of many rich ontologies) means
that many regularities are based on the same nesting of OWL constructors.
And a large branching factor signifies
that there is a good amount of variablitiy in term of the nesting of OWL constructors on some nesting level.

\begin{table}[t]
    \centering
    \caption{Depth and maximal branching factor of Hasse diagrams for posets.}
\label{table:hierarchyDepthBranching}

\resizebox{\textwidth}{!}{%
\begin{tabular}{| r | r | r | r || r | r | r | r | }
\hline
\multirow{2}{*}{\textbf{Depth}} & \multicolumn{3}{c||}{Axioms} & \multicolumn{3}{c|}{Class Frames}\\
\cline{2-7}
                                 & Atomic & $\mathcal{EL}^{++}$ & Rich & Atomic & $\mathcal{EL}^{++}$ & Rich \\
 \hline
 1        & 94 & 21 & 96  & 54 & 8  & 47\\
 2        & -  & 67 & 136 & 21 & 14 & 53\\
 3        & -  & 2  & 56  & 9  & 15 & 71\\
 4--9      & -  & -  & 139 & 10 & 33 & 212\\
 10--19    & -  & -  & 41  & -  & 17 & 61\\
 $\ge$ 20 & -  & -  & 5   & -  & 3  & 29\\
 \hline
\end{tabular} 
    \hspace*{0.1cm} 
\begin{tabular}{| r | r | r | r || r | r | r | r | }
\hline
\multirow{2}{*}{\textbf{Branching}} & \multicolumn{3}{c||}{Axioms} & \multicolumn{3}{c|}{Class Frames}\\
\cline{2-7}
                                 & Atomic & $\mathcal{EL}^{++}$ & Rich & Atomic & $\mathcal{EL}^{++}$ & Rich \\
 \hline
 0        & 94 & 21 & 96  & 54 & 8  & 47 \\
 1        & -  & 66 & 123 & 40 & 34 & 64 \\
 2        & -  & 2  & 55  & -  & 43 & 51 \\
 3--9      & -  & 1  & 174 & -  & 5  & 179\\
 10--19    & -  & -  & 25  & -  & -  & 70 \\
 $\ge$ 20 & -  & -  & -   & -  & -  & 62\\
 \hline
\end{tabular}}
\end{table}

\section{Related Work \& Discussion}
While there are many surveys of properties of existing ontologies, e.g.,
\cite{DBLP:conf/www/GlimmHKP12,DBLP:conf/semweb/MatentzogluBP13,DBLP:conf/ekaw/Svab-ZamazalS08,DBLP:conf/semweb/WangPH06},
there is only little research on the topic of discovering ontology patterns or reverse-engineering an ontology's design.
However, two approaches in this direction are motivated on similar grounds to the ones put forward in this work.

The first approach is based on agglomerative clustering to identify commonalities 
for named entities in an ontology based on similar syntactic representations
\cite{DBLP:conf/semweb/MikroyannidiISR11}.
Similarities between these representations are distilled in the form of sets of axioms with variables.
While these representations bear some similarities to the notion of modelling structures in the context of this work,
there are subtle differences with regards to the underlying notion of regularity. 
The approach using agglomerative clustering identifies regularities for named entities,
whereas the approach based on language abstractions identifies regularities for axioms (or sets of axioms).
So, the former approach is primarily concerned with regularities for elements of an ontology's domain-specific vocabulary,
whereas the latter focuses on regularities for syntactic structures based on an ontology's underlying formal language, e.g, OWL.

The second approach is based on frequent subtree mining over OWL axioms \cite{DBLP:journals/semweb/LawrynowiczPRT18}.
By interpreting OWL axioms as syntax trees, well-known subtree mining algorithms can be used to identify frequent tree structures.
Furthermore, a notion for regularities for class frames is motivated
that is based on identified regularities for syntax trees of axioms. 
For example, regularities for subsumption axioms with the same and non-variable left-hand side are grouped into a set 
to give rise to a new regularity for sets.
In cases where the left-hand side is a variable,
frequent itemset mining is proposed to identify co-occurring axioms as regularities for class frames.  
While the approach based on frequent subtree mining bears a resemblance to the approach based language abstractions,
there are both technical differences as well as conceptual differences. 

First and foremost, it is important to recognise
that frequent subtree mining aims at identify regularities based on some notion of \textit{frequency}.
A tree structure is considered frequent if it satisfies some threshold criterion.
However, regularities based on language abstractions are \textit{independent} of any notion of frequency; or any other notion depending on a threshold for that matter.
The importance of this needs to be emphasised because regularities based on thresholds
are generally not suitable for analysing an ontology's design as a whole.
The simple reason for this is that such notions, by definition, do not account for structures that do not satisfy the threshold criterion.  
For example, variations in the reuse of a single pattern in an ontology's design may give rise to many slightly different syntactic structures.
If none of the variant reuses of the pattern gives rise to frequent structures, then no regularity (based on frequency) is identified.  

In any case, any conclusion or claim about an ontology’s underlying design
based on syntactic regularities has to be made with due diligence regardless of the used approach.
Consider for example the case of a pattern-based ontology design.
A \textit{pattern} in the context of ontology engineering often denotes a rather distinctive notion.
An example of this are \textit{Ontology Design Patterns} (ODP)
that are proposed as well-proven modelling solution to common modelling problems
and often provide a reusable component such as a set of axioms \cite{DBLP:conf/semweb/Gangemi05,DBLP:conf/iceis/BlomqvistS05}.
While such a reusable component is often associated with a syntactic structure, e.g.,
a set of axioms, the converse is not necessarily the case.
Meaning, a reusable component of a pattern cannot be equated with the pattern itself
and the presence of axioms associated with a pattern's reusable component cannot
be equated with an actual reuse of the pattern.
So, even though the discovery of regularities can be helpful to detect structures 
that are indicative of an ODP's reuse,
a domain expert’s assessment of an identified regularity in an ontology is
required to gauge whether the regularity is connected to an ODP.

Even though the idea of reusable components has been popularised by the ODP community, 
there is no standard mechanism or de facto practice for reusing a given ODP.
Despite the development of frameworks and tool support for ODPs reuse 
\cite{DBLP:conf/esws/IannoneRS09,DBLP:conf/semweb/NoppensL09,DBLP:conf/semweb/SkjaevelandLKF18,DBLP:conf/swat4ls/WarrenderL13},
little is known about what kind of features are needed 
to facilitate pattern-based ontology engineering in practice \cite{DBLP:books/ios/p/HammarBCEFGHHJKKNSSS16}.
Developing an understanding of compositional aspects of syntactic structures in ontologies  
w.r.t.\ syntactic abstractions may provide a way of informing and evaluating the design of tools and frameworks in this direction.

As an example, consider the Galen Ontology \cite{DBLP:conf/rweb/RectorR06} in which the classes
$\conc{Current}\-\conc{Blood}\-\conc{Pressure}\-\conc{Level}$ and $\conc{RecentBloodPressureLevel}$
are represented via almost identical \opr{EquivalentClasses} axioms. Both use the following expression (written in infix notation):
\begin{center}
\resizebox{\textwidth}{!}{%
$\begin{array}[t]{lll}
     \conc{LevelState}~\sqcap~(\E \conc{isSpecificAnswerOf}.(\conc{InvestigationAct}~\sqcap (\E \conc{hasTimeOfOccurrence}.\\
     (\conc{TimeOfOccurrence}~\sqcap (\E \conc{hasAbsoluteState}~atTime))) \sqcap (\E \conc{isToDetermine}. \conc{BloodPressure})))
 \end{array}$ }
 \end{center}
 where the variable $atTime$ is set to $\conc{Now}$ and $\conc{RecentPast}$ respectively.
 Here, the use of the variable $atTime$ can be seen as an abstraction over differences between the representations of 
$\conc{Current}\-\conc{Blood}\-\conc{Pressure}\-\conc{Level}$ and $\conc{RecentBloodPressureLevel}$.
In this case, a simple templating mechanism allowing for the \textit{instantiation of parametrised representations},
e.g. $\conc{CurrentBloodPressureLevel} \equiv \texttt{BloodPressureLevel}(\conc{Now})$,
would be suitable to capture this abstract structure in an arguably meaningful way. 
So, research into the discovery of meaningful abstractions
as well as suitable ways of encoding them 
promises to have a great impact on pattern-based ontology engineering.

\section{Conclusion} 
In this paper, we adapted and extended a formal framework for analysing syntactic regularities in ontologies originally proposed in
\cite{DBLP:phd/ethos/Kindermann22,kindermann2021syntactic}.
The framework is based on a syntax-directed approach
that decomposes an ontology into equivalence classes of syntactic structures,
where two syntactic structures are considered equivalent
if they are indistinguishable under a formal notion of abstraction.  
We proposed the notion of a modelling structure for the purpose of analysing and characterising syntactic regularities.
Furthermore, we proposed formal relations between such modelling structures
so that they can be organised in terms of a partial order
that captures a notion of substructure containment.
Finally, we used these notions to conduct a large-scale empirical investigation of syntactic modelling structures in biomedical ontologies.

We find that most ontologies contain primarily axioms of a simple syntactic structure.
However, such axioms seem to be combined in various ways to give rise to comparatively many modelling structures for class frames. 
This suggests that class frames play a crucial role in the representation of many entities in the biomedical domain.

Our findings on common modelling structures across biomedical ontologies reveal 
that only comparatively simple syntactic structures for both axioms and class frames reoccur.
However, the results obtained on the maximal size and depth of modelling structures 
indicate that many rich ontologies also contain highly complex modelling structures in which OWL constructors are deeply nested and combined.
Moreover, such complex structures are also highly interrelated w.r.t.\ shared substructures in many ontologies.
While our investigation provides proof of structural complexities in ontologies,
further research is needed to qualify underlying design rationales.

\paragraph*{Supplemental Material Statement:} Source code is available at\\ \url{https://github.com/ckindermann/iswc-2022}.
\newpage 

 \bibliographystyle{splncs04}
 \bibliography{mybibliography}

\begin{thebibliography}{10}
\providecommand{\url}[1]{\texttt{#1}}
\providecommand{\urlprefix}{URL }
\providecommand{\doi}[1]{https://doi.org/#1}

\bibitem{DBLP:conf/dlog/2003handbook}
Baader, F., Calvanese, D., McGuinness, D.L., Nardi, D., Patel{-}Schneider, P.F.
  (eds.): The Description Logic Handbook: Theory, Implementation, and
  Applications. Cambridge University Press (2003)

\bibitem{DBLP:conf/iceis/BlomqvistS05}
Blomqvist, E., Sandkuhl, K.: Patterns in ontology engineering: Classification
  of ontology patterns. In: {ICEIS} {(3)}. pp. 413--416 (2005)

\bibitem{DBLP:conf/semweb/Gangemi05}
Gangemi, A.: Ontology design patterns for semantic web content. In: {ISWC}.
  Lecture Notes in Computer Science, vol.~3729, pp. 262--276. Springer (2005)

\bibitem{DBLP:conf/www/GlimmHKP12}
Glimm, B., Hogan, A., Kr{\"{o}}tzsch, M., Polleres, A.: {OWL:} {Y}et to arrive
  on the {W}eb of {D}ata? In: {LDOW}. {CEUR} Workshop Proceedings, vol.~937.
  CEUR-WS.org (2012)

\bibitem{DBLP:journals/ws/GrauHMPPS08}
Grau, B.C., Horrocks, I., Motik, B., Parsia, B., Patel{-}Schneider, P.F.,
  Sattler, U.: {OWL} 2: The next step for {OWL}. J. Web Semant.  \textbf{6}(4),
   309--322 (2008)

\bibitem{DBLP:books/ios/p/HammarBCEFGHHJKKNSSS16}
Hammar, K., Blomqvist, E., Carral, D., van Erp, M., Fokkens, A., Gangemi, A.,
  van Hage, W.R., Hitzler, P., Janowicz, K., Karima, N., Krisnadhi, A., Narock,
  T., Segers, R., Solanki, M., Sv{\'{a}}tek, V.: Collected research questions
  concerning ontology design patterns. In: Ontology Engineering with Ontology
  Design Patterns, Studies on the Semantic Web, vol.~25, pp. 189--198. {IOS}
  Press (2016)

\bibitem{DBLP:conf/owled/HorridgeP08}
Horridge, M., Patel{-}Schneider, P.F.: Manchester syntax for {OWL} 1.1. In:
  {OWLED} (Spring). {CEUR} Workshop Proceedings, vol.~496. CEUR-WS.org (2008)

\bibitem{DBLP:conf/esws/IannoneRS09}
Iannone, L., Rector, A.L., Stevens, R.: Embedding knowledge patterns into
  {OWL}. In: {ESWC}. Lecture Notes in Computer Science, vol.~5554, pp.
  218--232. Springer (2009)

\bibitem{DBLP:phd/ethos/Kindermann22}
Kindermann, C.: Analysing {P}atterns and {R}egularities in {O}ntologies. Ph.D.
  thesis, University of Manchester, {UK} (2022)

\bibitem{kindermann2021syntactic}
Kindermann, C., Parsia, B., Sattler, U.: Syntactic regularities based on
  language abstractions. Advances in Pattern-Based Ontology Engineering
  \textbf{51}, ~312 (2021)

\bibitem{DBLP:journals/semweb/LawrynowiczPRT18}
Lawrynowicz, A., Potoniec, J., Robaczyk, M., Tudorache, T.: Discovery of
  emerging design patterns in ontologies using tree mining. Semantic Web
  \textbf{9}(4),  517--544 (2018). \doi{10.3233/SW-170280},
  \url{https://doi.org/10.3233/SW-170280}

\bibitem{DBLP:conf/semweb/MatentzogluBP13}
Matentzoglu, N., Bail, S., Parsia, B.: A snapshot of the {OWL} web. In: {ISWC}
  {(1)}. Lecture Notes in Computer Science, vol.~8218, pp. 331--346. Springer
  (2013)

\bibitem{matentzoglu_nicolas_2017_439510}
Matentzoglu, N., Parsia, B.: Bioportal snapshot 30.03.2017 (Mar 2017).
  \doi{10.5281/zenodo.439510}, \url{https://doi.org/10.5281/zenodo.439510}

\bibitem{DBLP:conf/semweb/MikroyannidiISR11}
Mikroyannidi, E., Iannone, L., Stevens, R., Rector, A.L.: Inspecting
  regularities in ontology design using clustering. In: {ISWC} {(1)}. Lecture
  Notes in Computer Science, vol.~7031, pp. 438--453. Springer (2011)

\bibitem{motik2008}
Motik, B., Patel-Schneider, P., Bock, C., Fokoue, A., Haase, P., Hoekstra, R.,
  Horrocks, I., Ruttenberg, A., Sattler, U., Smith, M.: {OWL} 2 {W}eb
  {O}ntology {L}anguage: {S}tructural {S}pecification and {F}unctional-{S}tyle
  (01 2008), {WC}3 {R}ecommendation

\bibitem{DBLP:conf/semweb/NoppensL09}
Noppens, O., Liebig, T.: Ontology patterns and beyond - towards a universal
  pattern language. In: {WOP}. {CEUR} Workshop Proceedings, vol.~516.
  CEUR-WS.org (2009)

\bibitem{DBLP:journals/dke/NoyMMR04}
Noy, N.F., Musen, M.A., Jr., J.L.V.M., Rosse, C.: Pushing the envelope:
  challenges in a frame-based representation of human anatomy. Data Knowl. Eng.
   \textbf{48}(3),  335--359 (2004)

\bibitem{DBLP:journals/biomedsem/Osumi-Sutherland17}
Osumi{-}Sutherland, D., Courtot, M., Balhoff, J.P., Mungall, C.J.: Dead simple
  {OWL} design patterns. J. Biomed. Semant.  \textbf{8}(1),  18:1--18:7 (2017)

\bibitem{DBLP:conf/rweb/RectorR06}
Rector, A.L., Rogers, J.: Ontological and practical issues in using a
  description logic to represent medical concept systems: Experience from
  {GALEN}. In: Reasoning Web. Lecture Notes in Computer Science, vol.~4126, pp.
  197--231. Springer (2006)

\bibitem{DBLP:journals/biomedsem/SarntivijaiLXMDVSPMPLTSMNBHZSPMCSAH14}
Sarntivijai, S., Lin, Y., Xiang, Z., Meehan, T.F., Diehl, A.D., Vempati, U.D.,
  Sch{\"{u}}rer, S.C., Pang, C., Malone, J., Parkinson, H.E., Liu, Y.,
  Takatsuki, T., Saijo, K., Masuya, H., Nakamura, Y., Brush, M.H., Haendel,
  M.A., Zheng, J., Jr., C.J.S., Peters, B., Mungall, C.J., Carey, T.E., States,
  D.J., Athey, B.D., He, Y.: {CLO:} the cell line ontology. J. Biomed. Semant.
  \textbf{5}, ~37 (2014)

\bibitem{DBLP:conf/semweb/SkjaevelandLKF18}
Skj{\ae}veland, M.G., Lupp, D.P., Karlsen, L.H., Forssell, H.: Practical
  ontology pattern instantiation, discovery, and maintenance with reasonable
  ontology templates. In: {ISWC} {(1)}. Lecture Notes in Computer Science, vol.
  11136, pp. 477--494. Springer (2018)

\bibitem{DBLP:conf/ekaw/Svab-ZamazalS08}
Sv{\'{a}}b{-}Zamazal, O., Sv{\'{a}}tek, V.: Analysing ontological structures
  through name pattern tracking. In: {EKAW}. Lecture Notes in Computer Science,
  vol.~5268, pp. 213--228. Springer (2008)

\bibitem{DBLP:conf/semweb/WangPH06}
Wang, T.D., Parsia, B., Hendler, J.A.: A survey of the web ontology landscape.
  In: {ISWC}. Lecture Notes in Computer Science, vol.~4273, pp. 682--694.
  Springer (2006)

\bibitem{DBLP:conf/swat4ls/WarrenderL13}
Warrender, J.D., Lord, P.: A pattern-driven approach to biomedical ontology
  engineering. In: {SWAT4LS}. {CEUR} Workshop Proceedings, vol.~1114.
  CEUR-WS.org (2013)

\end{thebibliography}
%
%
%
%
%
\end{document}